\documentclass[letterpaper]{article} 
\usepackage{aaai2026}  
\usepackage{times}  
\usepackage{helvet}  
\usepackage{courier}  
\usepackage[hyphens]{url}  
\usepackage{graphicx} 
\usepackage{natbib}  
\usepackage{caption} 
\usepackage{amsmath}
\usepackage{cleveref}
\crefname{figure}{Fig.}{Figs.}
\crefname{figure}{Figure}{Figures}
\crefname{section}{Sec.}{Secs.}
\Crefname{section}{Section}{Sections}
\Crefname{table}{Table}{Tables}
\crefname{table}{Tab.}{Tabs.}
\usepackage{algorithm}
\usepackage{algorithmic}

\usepackage{multirow,adjustbox}
\usepackage{amssymb}
\usepackage{pifont}
\usepackage{xcolor,colortbl}

\definecolor{Gray}{gray}{0.85}
\definecolor{mygreen}{rgb}{0,.7,0}
\newcolumntype{g}{>{\columncolor{Gray}}c}
\newcommand{\cmark}{\textcolor{mygreen}{\ding{51}}}
\newcommand{\xmark}{\textcolor{red}{\ding{55}}}
\newcommand{\name}{\textsc{trust}}

\usepackage{newfloat}
\usepackage{listings}
\DeclareCaptionStyle{ruled}{labelfont=normalfont,labelsep=colon,strut=off} 
\floatstyle{ruled}
\newfloat{listing}{tb}{lst}{}
\floatname{listing}{Listing}
\title{TRUST: Leveraging Text Robustness for Unsupervised Domain Adaptation}

\author{
    Mattia Litrico\textsuperscript{\rm 1}, Mario Valerio Giuffrida\textsuperscript{\rm 2}, Sebastiano Battiato\textsuperscript{\rm 1}, Devis Tuia\textsuperscript{\rm 3}
}
\affiliations{
    \textsuperscript{\rm 1}University of Catania, Italy\\
    \textsuperscript{\rm 2}University of Nottingham, UK\\
    \textsuperscript{\rm 3}École Polytechnique Fédérale de Lausanne (EPFL), Switzerland\\

}

\usepackage{bibentry}
\usepackage{booktabs}

\begin{document}

\maketitle

\begin{abstract}
Recent unsupervised domain adaptation (UDA) methods have shown great success in addressing classical domain shifts (e.g., synthetic-to-real), but they still suffer under complex shifts (e.g. geographical shift), where both the background and object appearances differ significantly across domains. Prior works showed that the language modality can help in the adaptation process, exhibiting more robustness to such complex shifts. In this paper, we introduce \name, a novel UDA approach that exploits the robustness of the language modality to guide the adaptation of a vision model. \name~generates pseudo-labels for target samples from their captions and introduces a novel uncertainty estimation strategy that uses normalised CLIP similarity scores to estimate the uncertainty of the generated pseudo-labels. Such estimated uncertainty is then used to reweight the classification loss, mitigating the adverse effects of wrong pseudo-labels obtained from low-quality captions. To further increase the robustness of the vision model, we propose a multimodal soft-contrastive learning loss that aligns the vision and language feature spaces, by leveraging captions to guide the contrastive training of the vision model on target images. In our contrastive loss, each pair of images acts as both a positive and a negative pair and their feature representations are attracted and repulsed with a strength proportional to the similarity of their captions. This solution avoids the need for hardly determining positive and negative pairs, which is critical in the UDA setting. Our approach outperforms previous methods, setting the new state-of-the-art on classical (DomainNet) and complex (GeoNet) domain shifts. The code will be available upon acceptance.
\end{abstract}


\begin{figure}[t!]
    \centering
    \includegraphics[trim={0 0 2.45cm 0},clip,width=0.8\linewidth]{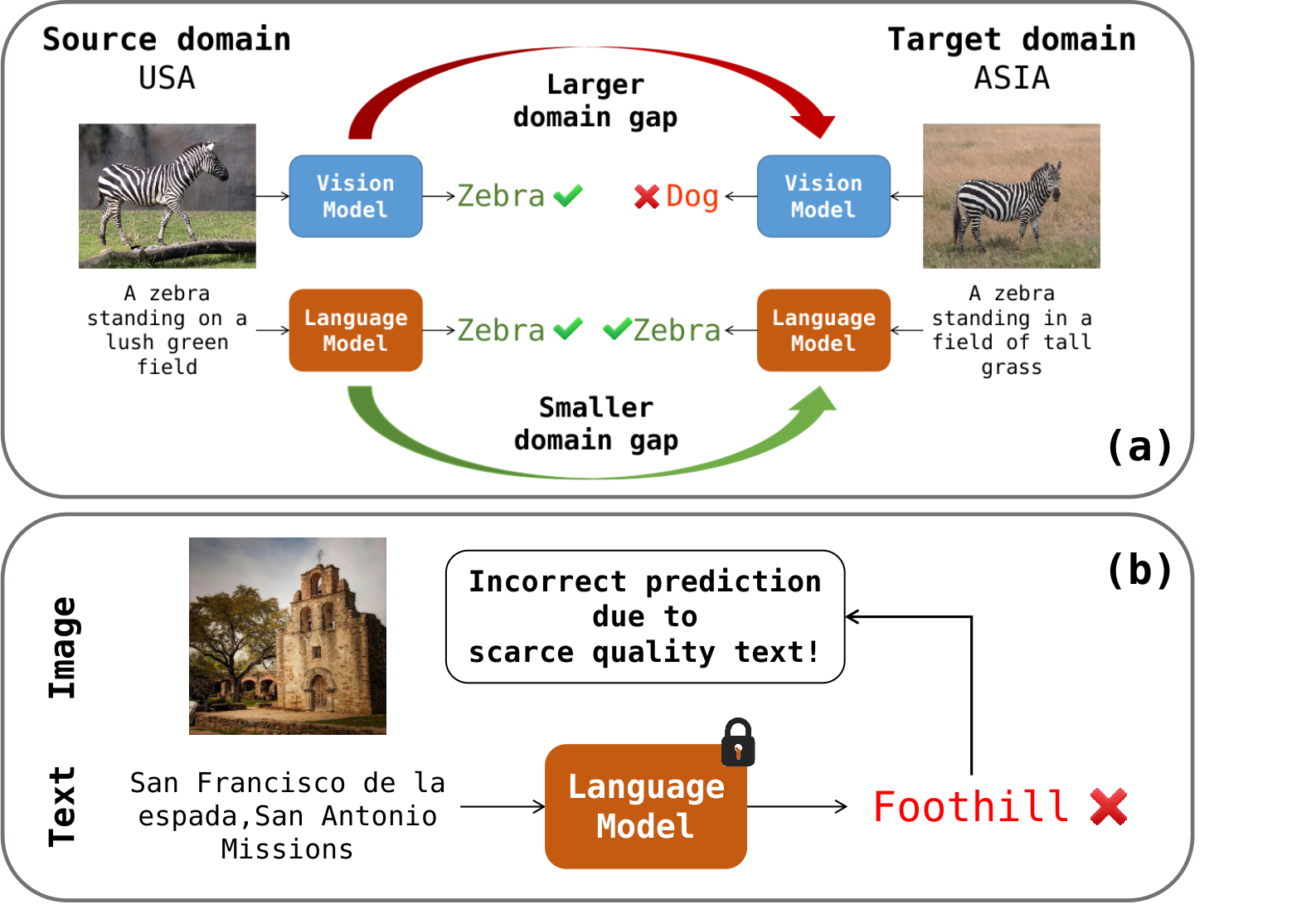}
    \caption{(a) geographical shifts strongly impact the appearance of both foreground objects and background.
    Contrarily, captions contain valuable information about the semantic class and they are minimally affected by the geographic shift, as stated in \cite{lagtran}. (b) Despite such apparent robustness, low quality captions may still lead to low classification accuracy.}
    \label{fig:figure_1}
    \vspace{-1em}
\end{figure}

\section{Introduction}
\label{sec:intro}
Although deep learning models have achieved remarkable performance in lots of computer vision tasks, they still fall short in their ability to generalise to different domains.
Retraining deep learning models on new data requires a big effort to acquire and manually label images and should be avoided.
To address these limitations, Unsupervised Domain Adaptation (UDA) approaches for image classification have been proposed. UDA aims at transferring knowledge acquired on a labelled source domain to an unlabelled target domain, bridging the gap between them  \cite{uda3,toalign,uda4,daln}.
Recent UDA methodologies have been successful on classical domain shifts (\textit{e.g.} synthetic-to-real), but they suffer a lot on more complex shifts (\textit{e.g.} geographical), where both the background and objects' appearance change between domains \cite{geonet}.

Kalluri et al. \cite{lagtran} posited that, under complex shift, solely relying on images is challenging and they proposed LaGTran that integrates textual data for guiding the adaptation. Indeed, authors showed that textual data are more robust to complex shifts, as they semantically describe the image content instead of focusing on appearance details, which is essential for bridging the domain gap (\textit{c.f.} \cref{fig:figure_1}(a)).
However, LaGTran \cite{lagtran} used textual data only for generating pseudo-labels for the target domain, highly underusing the potential of the language modality to reduce the domain shift. Moreover, they blindly rely on the pseudo-labels generated from the text, which may be incorrect due to two main factors: (a) the scarce quality of text descriptions, especially for crowd-sourced texts (\textit{e.g.} \cref{fig:figure_1} (b)); (b) the impact of domain shifts on language models, which is lower than image-based models but still present.
For those reasons, solely training the vision model with pseudo-labels generated from the text is suboptimal for transferring the shift robustness from the language to the vision model.

To overcome these limitations, we propose \textbf{\name} (\textbf{T}ext \textbf{R}ob\textbf{UST}ness for unsupervised domain adaptation), a novel approach for UDA in image classification that exploits the potential of language guidance for the adaptation of a vision model. 
Similarly to LaGTran \cite{lagtran}, we use a language model to generate pseudo-labels on target samples from captions, but we also propose two novel components to cope with the shortcomings discussed above. 

First, to reduce the impact of incorrect pseudo-labels generated from low-quality captions, we propose a novel multimodal uncertainty estimation strategy that reweights the classification loss for target samples, by evaluating how well the captions semantically describe the target images. 
To this aim, we use a pretrained vision-language model (\textit{i.e.} CLIP \cite{clip}) to evaluate the semantic correlation between target images and their captions, which serves as a measure of the uncertainty/reliability of generated pseudo-labels. 
The estimated uncertainty is then used to reweight the contribution of the pseudo-labels in the classification loss, reducing the negative impact of wrong pseudo-labels obtained from low-quality captions.

Second, to enhance the generalisation of the vision model, we aim to transfer the robustness to complex shifts from the language to the vision model, by promoting an effective alignment between their feature spaces. We propose to integrate a novel multimodal soft-contrastive learning loss, which uses the language modality to guide the contrastive training of the vision model. Unlike previous works, our soft-contrastive framework does not require to identify positive and negative pairs, as it assigns to each pair a score of ``\textit{positiveness}'' and ``\textit{negativeness}'' based on how likely they share the same semantic content. Then, a pair of images acts simultaneously as a positive and negative pair and attracts/repulses samples with a strength proportional to the similarity of their captions. The benefits of this strategy are multiple: (a) we encourage the vision model to match the language model's feature space by attracting representations of images with similar captions and repulsing those with dissimilar ones; (b) differently than \citet{adacontrast,litrico_2023_CVPR}, we avoid contrasting images of the same class without relying on the output of the vision model, therefore limiting the confirmation bias; (c) we achieve a smoother contrastive training, where representations of each pair of samples are both attracted and repulsed based on how likely they share the same semantic content.

We benchmark \name~on three datasets representing classical (DomainNet, VisDA) and complex (GeoNet) shifts. On GeoNet, we obtain the best performance with a gain of $+2.96\%$ to the best competitor; on DomainNet and VisDA, we improve performance of $+2.53\%$ and $+2.10\%$, compared to state-of-the-art. 

To summarise, our main contributions are:
\begin{itemize}
    \item We introduce a novel uncertainty estimation strategy that leverages CLIP to evaluate the uncertainty of pseudo-labels, by measuring the semantic correlation between images and their captions. 
    \item We propose a novel multimodal soft-contrastive learning loss that uses the language modality to guide the contrastive training of the vision model. Our solution removes the problem of identifying positive and negative samples in UDA, preventing the confirmation bias.
    \item We validate our method on classical and complex domain shifts outperforming the state-of-the-art on both settings.
\end{itemize}

\section{Related Works}
\label{sec:related_works}

\begin{figure*}[t]
    \centering
    \includegraphics[width=0.8\linewidth]{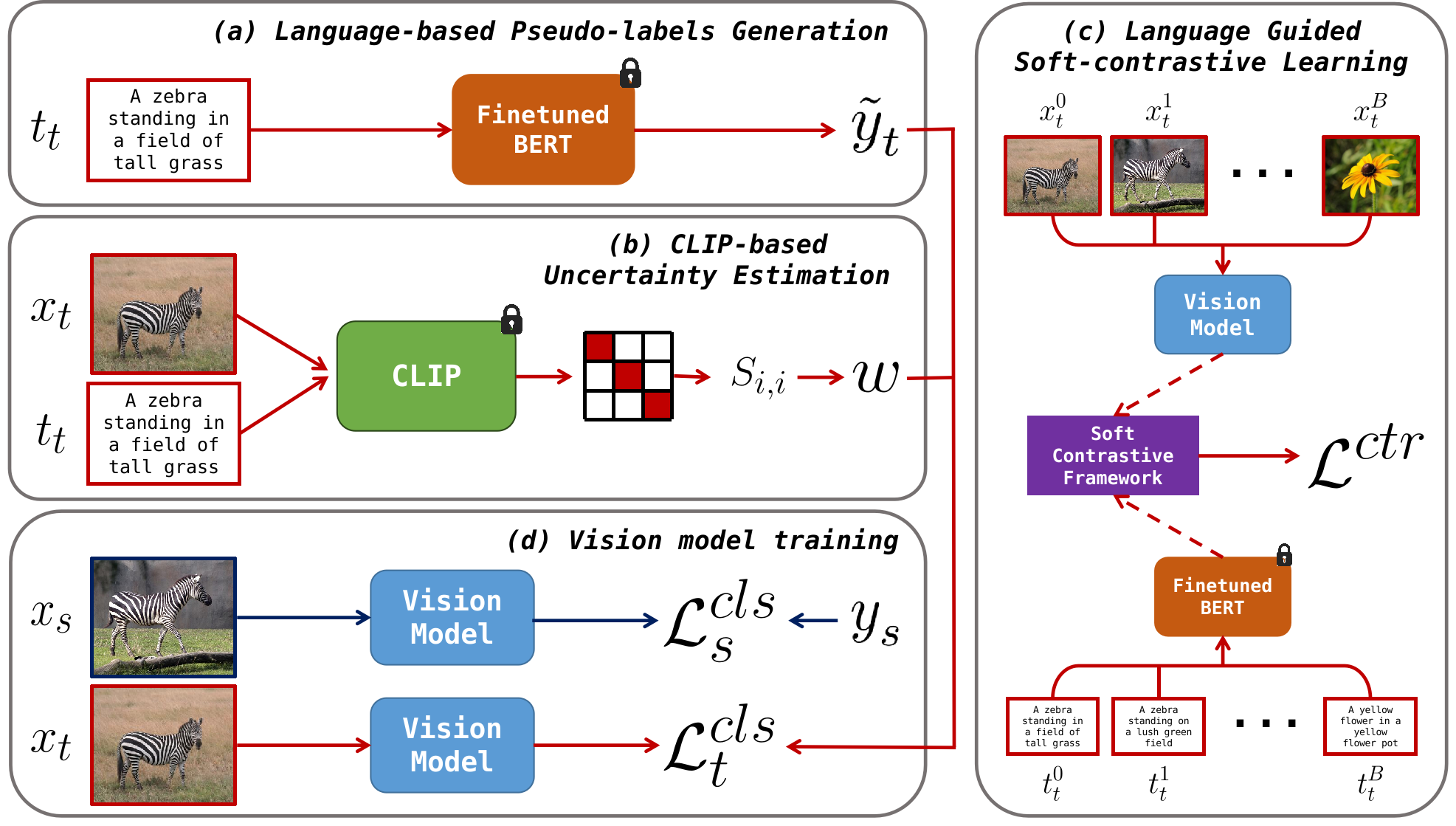}
    \caption{(a) We use the finetuned BERT model on target captions to generate pseudo-labels $\mathbf{\tilde{y}_t}$ for target samples (\cref{subsec:lagtran}). (b) From target images and texts, we compute normalised CLIP similarity scores $s^{i,i}$ used to estimate $\mathbf{w}$ as a measure of the reliability of the generated pseudo-labels $\mathbf{\tilde{y}_t}$ (\cref{subsec:uncertainty}). (c) Feature representations of images go through a soft-contrastive framework, where they are attracted and repulsed to each other based on the similarity of their captions (dashed lines indicates features extraction). (d) A vision model is trained on both source and target images. On source images, we compute a classification loss $\mathcal{L}_s^{cls}$ using the ground truth labels $\mathbf{y}_s$. On target images we use the pseudo-labels $\mathbf{\tilde{y}_t}$ as supervision and the classification loss $\mathcal{L}_t^{cls}$ is reweighted based on the estimated reliability score $\mathbf{w}$.}
    \label{fig:main_fig}
    \vspace{-1em}
\end{figure*}

\noindent \textbf{Unsupervised Domain Adaptation}. Unsupervised Domain Adaptation aims to adapt a model trained on a labelled source domain to generalise well on an unlabelled target domain, in presence of domain shift \cite{uda1}. 
Earlier UDA approaches focused on aligning statistical distributions \cite{domainnet}. More recent methods proposed domain alignment strategies, including Maximum Mean Discrepancy \cite{domain_alignment_1,domain_alignment_2}, adversarial learning \cite{toalign,daln}, clustering \cite{clustering_da,clustering_da_2} and self-training \cite{adacontrast,litrico_2023_CVPR}. Other works leveraged generative models \cite{generative_da} and transformers \cite{pmtrans,transformer_da} to learn the underlying distributions of data using the generation process or the attention mechanism. However, all of these approaches solely operate on the image space, which has been demonstrated to be suboptimal for complex shifts \cite{geonet}. Differently, \citet{llm_da} proposed to integrate a large language model (LLM) for UDA, while other works \cite{language_da,language_da2,language_da3,language_da4} leveraged the language modality for domain generalisation, but they rely on short class descriptors, which are not semantically rich as crowd-sourced text.
For example, \citet{goyal} used class names as text descriptors, which are less semantically informative than free-form texts.
To overcome these limitations, LaGTran \cite{lagtran} used captions to generate pseudo-labels as a source of supervision for target samples. However, given the uncurated nature of captions, the generated supervision may be incorrect, often leading to overfitting the introduced noise.

\noindent \textbf{VLMs in Knowledge Transfer}.
Vision-Language Models, such as CLIP \cite{clip} and ALIGN \cite{align}, demonstrated success in capturing modality-invariant features, opening new avenues for knowledge transfer. Some works \cite{padclip,damp,unimos} leveraged the zero-shot ability of VLMs, applying or finetuning them to the target domain. In \cite{clip_da2}, the authors used prompt or adaptor learning to fine-tune the VLM to the target domain in a semi-supervised fashion. DAPL \cite{dapl} learned domain-specific prompts to separate domain and class information in the CLIP visual feature space. DIFO \cite{difo} used prompt learning to adapt the VLM to the target domain and distilled it to a target model.
In contrast, we propose a different paradigm using CLIP to estimate the uncertainty of pseudo-labels generated by a language model, by leveraging the multimodal semantic alignment of CLIP's feature spaces. 

\noindent \textbf{Contrastive Learning in Domain Adaptation}.
Self-supervised methods proved their effectiveness in learning generalised representations of visual data \cite{simclr,selflearning1,selflearning2}.
Specifically, contrastive-based approaches \cite{simclr,selflearning3} have demonstrated to improve the generalisation of deep learning models.
In \cite{contrastive1,simclr}, the authors proposed a self-supervised contrastive framework that used augmentations to generate positive pairs and all the other training samples for the negative pairs. Other approaches \cite{contrastive2,contrastive3,contrastive4} proposed strategies to optimise the selection of negative samples to improve the contrastive training. However, these methods include in the negative pairs also samples sharing the same class, disrupting the adaptation. Recent approaches \cite{adacontrast,litrico_2023_CVPR} proposed to exclude samples of the same class from the negative pairs, by looking at pseudo-labels \cite{adacontrast}, historical predictions \cite{litrico_2023_CVPR}, or clustering assignments \cite{contrastive5}. Since this exclusion is based on the model predictions, it easily leads to confirmation bias.
Differently, our contrastive framework does not require the selection of positive and negative pairs, overcoming the aforementioned issues. Specifically, we propose a soft-contrastive loss, where each pair simultaneously acts as both a positive and negative pair. The representations of each pair are then attracted and repulsed based on an estimated score of ``\textit{positiveness}'' and ``\textit{negativeness}''.

\section{Proposed Method}
\label{sec:method}


An overview of \name~is shown in \cref{fig:main_fig}. Our aim is unsupervised domain adaptation (UDA), \textit{i.e.} to learn a model for the target domain without having access to any ground-truth labels from it. Here, we have a labelled source domain $\mathcal{D}_s : \{x_s^i, t_s^i, y_s^i \}^{N_s}_{i=1}$, where $x_s$ and $y_s$ are source images and ground-truth labels, and $t_s$ are source captions. Similarly, we have an unlabelled target domain $\mathcal{D}_t : \{x_t^i, t_t^i \}^{N_t}_{i=1}$. Captions are obtained from either associated metadata in web-collected images \cite{mahajan}, or generated with image-to-text models \cite{blip2} and they are used \textit{only} at training time. \name~trains a vision model composed of a feature extractor $f : \mathbb{R}^{H \times W \times 3} \rightarrow \mathbb{R}^P$ and a classifier $h : \mathbb{R}^P \rightarrow \mathbb{R}^C$ , where $P$ is the length of the feature vector, and $C$ is the number of classes.


\subsection{Language Guided Domain Adaptation}
\label{subsec:lagtran}
We address the problem of UDA leveraging pseudo-labels generated from captions. Similarly to \cite{lagtran}, we fine-tune a BERT sentence classifier \cite{bert} in a supervised fashion using captions and labels from the source domain. The trained BERT model is then frozen and inputted with target captions to generate pseudo-labels for target samples. Then, source labels and the obtained target pseudo-labels are used as supervision to train a vision model on both source and target domains simultaneously, as detailed in \cref{subsection:overall_framework}.

More formally, we fine-tune a pretrained DistilBERT \cite{distilBert} model $\mathcal{B}$ on the source domain $\{ t_s^i, y_s^i\}^{N_s}_{i=1}$ using source captions and labels, optimising this objective:

\begin{equation}
    \underset{\phi}{\operatorname{argmin}} \; \mathbb{E}_{\{t^i_s, y^i_s\} \sim \mathcal{D}_s} \; \mathcal{L}_{\operatorname{CE}} (\mathcal{B}_{\phi}(t_s^i), y_s^i),
\end{equation}
where $\phi$ denotes the parameters of the BERT model and $\mathcal{L}_{\operatorname{CE}}$ is the cross-entropy loss.
Then, we use the fine-tuned model $\mathcal{B}$ to generate pseudo-labels $\mathbf{\tilde{y}_t}$ on target captions $\mathbf{t_t}$ by running inference passes as follows:

\begin{equation}
    \tilde{y}^i_t = \underset{C}{\operatorname{argmax}} \; \mathcal{B}_{\phi}(t_t^i).
\end{equation}

\begin{figure}[t]
    \centering
    \includegraphics[width=0.8\linewidth]{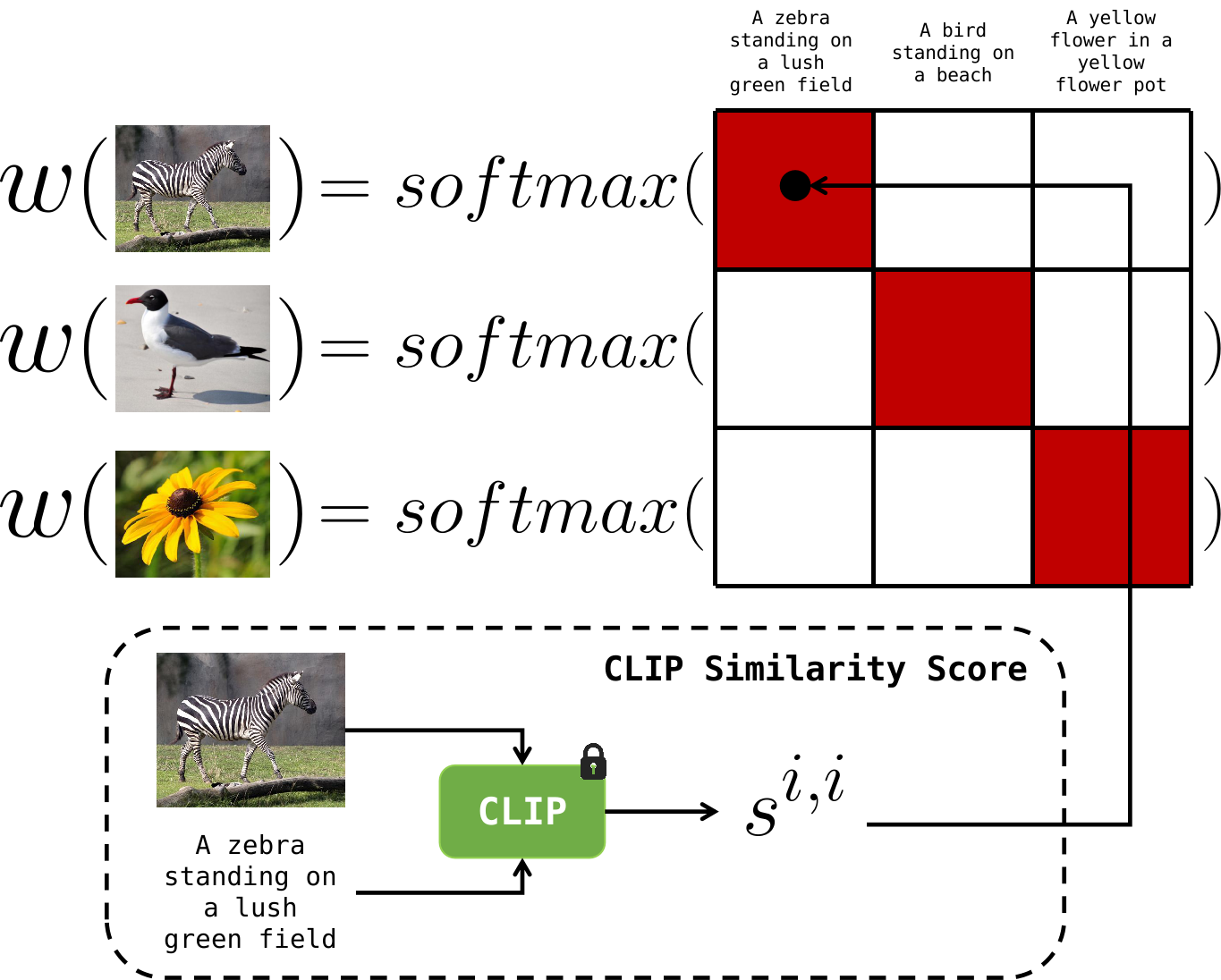}
    \caption{We compute CLIP similarity scores on each pair of target images and texts in a batch, obtaining the similarity matrix $S$. To calculate the reliability weight $\mathbf{w}$, we normalise the CLIP scores with a softmax on each row of $S$ and select the values in the main diagonal, which measure the semantic correlation between much each image and its caption.}
    \label{fig:clip_uncertainty}
    \vspace{-1em}
\end{figure}

\subsection{CLIP-based Pseudo-labels Uncertainty Estimation}
\label{subsec:uncertainty}
Although previous works \cite{lagtran} showed that the language modality is more robust to complex shifts, blindly relying on the knowledge acquired from captions may lead to the generation of incorrect pseudo-labels $\mathbf{\tilde{y}_t}$ due to: (a) the scarce quality of captions (\textit{c.f.} \cref{fig:figure_1}(b)), especially when they are crowd-sourced from the web; and (b) the domain shift that still exists in the language modality.
Therefore, training an image classifier on such pseudo-labels (as in \cite{lagtran}) risks to disrupt the adaptation process.

To mitigate these issues, we propose a novel strategy to estimate the uncertainty/reliability of the pseudo-labels generated from target captions, based on the capacity of captions to semantically describe the corresponding images. Such uncertainty scores are then used to reweight the classification loss, accordingly. With this aim, we use a pretrained CLIP model \cite{clip} to evaluate the correlation between images and corresponding captions. 
The underlying idea, shown in \cref{fig:clip_uncertainty}, is that when the CLIP image/text similarity is high, the text accurately describes the image content. Consequently, we will assume that the pseudo-label generated with BERT (\textit{c.f.} \cref{subsec:lagtran}) is reliable. On the contrary, when the CLIP similarity is low, the text does not describe well the content of the image (e.g. \cref{fig:figure_1} (b)). This may occur when the text is of limited quality or because it does not capture the semantics of the image. In this case, the pseudo-label is considered unreliable.
Differently from other uncertainty estimation strategies \cite{litrico_2023_CVPR}, which perform such estimation based on the output of the training model, our solution based on CLIP prevents the confirmation bias \cite{DivideMix}, since the CLIP estimation is not affected by the training of the \name's vision model, and CLIP's parameters are frozen during the adaptation process.

More formally, given a batch of image/caption pairs from the target domain, we compute the similarity matrix $S(E_{\operatorname{im}}(\mathbf{x_t}), E_{\operatorname{text}}(\mathbf{t_t}))$ as the cosine similarity between the embeddings of target images $\mathbf{x_t}$ and texts $\mathbf{t_t}$ obtained from the CLIP's image and text encoders $E_{\operatorname{im}}$ and $E_{\operatorname{text}}$. The element $s^{i,j}$ of the similarity matrix indicates the semantic correlation between the $i$-th image and the $j$-th caption in the batch. Consequently, the diagonal elements $s^{i,i}$ indicates how much each target image is semantically related to its caption. However, this score is unbounded and needs to be normalised to measure the reliability of the pseudo-labels obtained from the captions. Therefore, we compute the softmax function for each row of the similarity matrix $S$ to obtain a normalised score $w_i$, as follows:

\begin{equation}
\label{eq:clip_softmax}
    w_i = \frac{\exp({s^{i,i}})}{\sum_{j=1}^B \exp({s^{i,j}})}.
\end{equation}

The resulting score $w_i$ is an estimation of the quality of the caption $t_t^i$ to describe the image $x_t^i$.
We use the softmax function for normalising $s^{i,i}$, to produce a score proportional to how much each image is semantically related to its caption (with respect to the other texts in the batch). 
The larger $w_i$ (high reliability), the better the $i$-th image is described by its corresponding text, which leads to a higher confidence in the pseudo-label $\mathbf{\tilde{y}_i}$. Conversely, a lower value of $w_i$ (low reliability) means that the caption semantically describes the image as less as the other texts in the batch.
Note that CLIP is pretrained and frozen during this process, to avoid adding an additional overhead for finetuning CLIP and preventing the confirmation bias.

\subsection{Language Guided Soft-contrastive Learning}
\label{subsection:soft_contrastive}
The language modality intrinsically benefits of a larger robustness to complex domain shifts compared to visual data, as demonstrated in \cite{geonet}. 
We posit that combining the benefits of language and vision modalities improves the performance on the target domain.  
LaGTran \cite{lagtran} used a language model to generate pseudo-labels on target images for training a vision model. Despite its simplicity, this strategy alone does not encourage the vision model to inherit the robustness of the language model. 

Therefore, we propose a novel language guided soft-contrastive learning framework to transfer the intrinsic robustness from the language to the vision model, by aligning their feature spaces.
Differently than previous works \cite{moco,simclr,adacontrast}, our contrastive framework treats each pair of images as both a positive and a negative pair, with a score of ``\textit{positiveness}'' and ``\textit{negativeness}'' based on the semantic similarity between image captions. In this way, the feature representation of each pair will be both attracted and repulsed with a strength proportional to how likely they share the same semantic content. This strategy leads to multiple benefits.
Firstly, it aligns the language and vision feature spaces, transferring the intrinsic robustness to complex domain shifts from the language to the vision model. 
Secondly, it does not require to determine positive and negative pairs, since each pair plays simultaneously as a positive and a negative pair. This avoids using the vision model itself to discriminate between positive and negative pairs \cite{litrico_2023_CVPR}, reducing the effects of confirmation bias. 
Finally, our solution reduces the adverse effects of mistakenly assigning a pair as positive or negative. If two samples have incorrect pseudo-labels, the pair will not be strictly assigned to either the positive or negative class. Instead, their feature representations are both attracted and repulsed, limiting the impact of the incorrect assignment.

More formally, let $\mathcal{S} = \{ a_w(x_t^i), a_s(x_t^i) \}_{i=1}^B$ be a batch composed of target samples, where we apply a weak $a_w \in \mathcal{A}_w$ and a strong augmentation $a_s \in \mathcal{A}_s$ drawn from distributions $\mathcal{A}_w$ and $\mathcal{A}_s$.
Standard self-supervised contrastive methods \cite{simclr} optimise the following: 
\begin{equation}
\label{eq:self_ctr}
    \mathcal{L}^{ctr}_{self} = - \sum_{i}^B \log \frac{exp(z_i \cdot \bar{z}_i / \tau)} {\sum_{j \in \mathcal{N}} exp(z_i \cdot z_j / \tau)},
\end{equation}
where $z_i = f(a_w(x_t^i))$ and $\bar{z_i} = f(a_s(x_t^i))$ are feature representations extracted from the weakly and strongly augmented $i$-th target sample, $\tau$ is a temperature parameter, and $\mathcal{N}:\{ j | 1\leq j \leq B, j \neq i \}$ is the set of indices of all the negative samples.
In this formulation, only the augmented version of the same sample is treated as positive, while all the other samples are treated as negatives.

Inspired by \cite{supcontrast}, we generalise this formulation to account for multiple positive samples. Each pair of images in the batch is treated as positive, with a score of positiveness depending on how similar their captions are, leading to the following objective: 
\vspace{-1em}
\begin{multline}
\vspace{-1em}
    \mathcal{L}^{ctr} =\\
    - \sum_{i}^B \log \Biggl\{ \frac{1}{|\mathcal{S}|} \sum_{p=1}^{|\mathcal{S}|} \frac{ \operatorname{sim}_{(i,p)} \cdot \exp(z_i \cdot \bar{z}_p / \tau)} {\sum_{j \in \mathcal{N}} (1-\operatorname{sim}_{(i,j)}) \cdot \exp(z_i \cdot z_j / \tau)} \Biggl\},
\vspace{-1em}
\end{multline}

where $\operatorname{sim}_{(a,b)} = \frac{f^{\mathcal{B}} (t_t^a) \cdot f^{\mathcal{B}} (t_t^b)}{||f^{\mathcal{B}} (t_t^a)|| \cdot ||f^{\mathcal{B}} (t_t^b)||}$ is the cosine similarity between the DistilBERT features vectors for the text descriptions of target samples $x_t^a$ and $x_t^b$.
Differently than \cref{eq:self_ctr}, we treat each sample in the batch (e.g. $\bar{z}_p$) as positive of the $i$-th sample, requiring to include another summation over the cardinality of the multiviewed batch. This results in attracting the feature representations of a pair of images $(x_t^a, x_t^b)$ with a strength equal to $\operatorname{sim}_{(a,b)}$, while repulsing them with a strength equal to $1-\operatorname{sim}_{(a,b)}$.

\begin{figure}[t]
    \centering
     \includegraphics[width=0.7\linewidth]{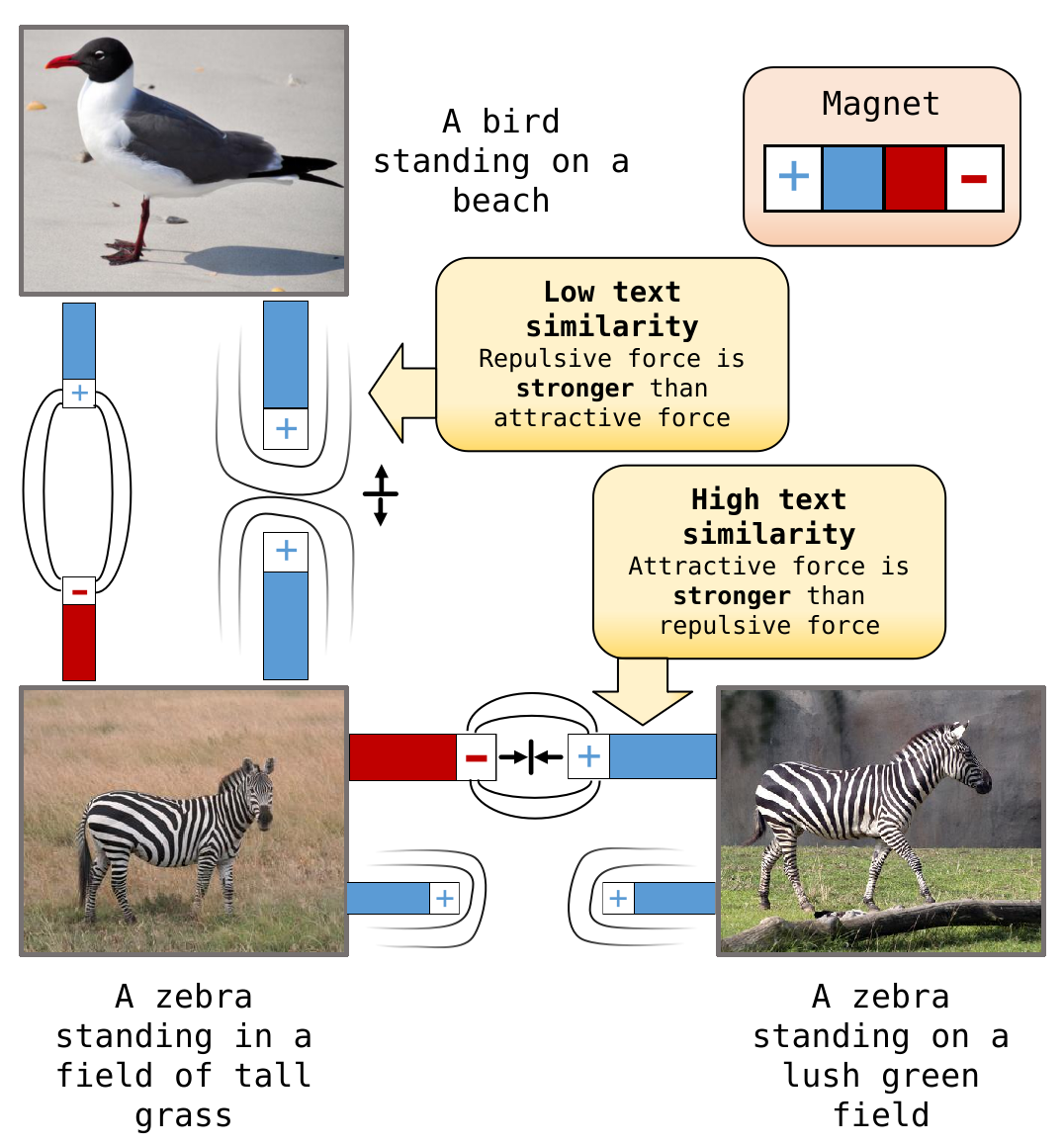}
    \caption{Representation of our soft-contrastive framework, where each pair of images is both attracted and repulsed based on their caption similarity: if two images have semantically similar captions, they are more attracted than repulsed and vice versa. Attraction/Repulsion is shown by magnets with opposite/same polarity. The strength of each force is indicated by magnet size and field lines.}
    \label{fig:soft_contrastive}
    \vspace{-1em}
\end{figure}

\begin{table*}
    \centering
    \begin{adjustbox}{width=0.9\textwidth}
    \begin{tabular}{lcccccccccccccccc}
        \toprule
         &  \multicolumn{3}{c}{Real $\rightarrow$}  & & \multicolumn{3}{c}{Clipart $\rightarrow$} & & \multicolumn{3}{c}{Sketch $\rightarrow$} & & \multicolumn{3}{c}{Painting $\rightarrow$} & \multirow{ 2}{*}{Avg.}\\
         \cline{2-4} \cline{6-8} \cline{10-12} \cline{14-16}
         & C & S & P & & R & S & P & & R & C & P & & R & C & S\\
         
          \midrule
         \multicolumn{14}{l}{\textit{\textbf{Zero-shot Classification}}}\\ 
         CLIP \cite{clip} & 72.39 & 60.90 & 66.81 & & 81.37 & 60.90 & 66.81 & & 81.37 & 72.39 & 66.81 & & 81.37 & 72.39 & 60.90 & 70.38 \\
         CLIP* \cite{clip} & 73.45 & 62.81 & 67.07 & & \textbf{81.71} & 62.81 & \underline{67.07} & & \underline{81.71} & 73.45 & 67.07 & & \underline{81.71} & 73.45 & 62.81 & 71.26 \\ 
         \midrule
         \multicolumn{14}{l}{\textit{\textbf{UDA}}}  \\
         Source only &  63.02 & 49.47 & 60.48 & & 70.52 & 56.09 & 52.53 & & 70.42 & 65.91 & 54.47 & & 73.34 & 60.09 & 48.25 & 60.38 \\
         MCD \cite{uda2} & 39.40 & 25.20 & 41.20 & & 44.60 & 31.20 & 25.50 & & 34.50 & 37.30 & 27.20 & & 48.10 & 31.10 & 22.80 & 34.01 \\
         MDD \cite{mdd} &  52.80 & 41.20 & 47.80 & & 52.50 & 42.10 & 40.70 & & 54.20 & 54.30 & 43.10 & & 51.20 & 43.70 & 41.70 & 47.11 \\
         CGDM \cite{cgdm} & 49.40 & 38.20 & 47.20 & & 53.50 & 36.90 & 35.30 & & 55.60 & 50.10 & 43.70 & & 59.40 & 37.70 & 33.50 & 45.04 \\
         SCDA \cite{clustering_da_2} & 54.00 & 42.50 & 51.90 & & 55.00 & 44.10 & 39.30 & & 53.20 & 55.60 & 44.70 & & 56.20 & 44.10 & 42.00 & 48.55 \\
         SSRT-B \cite{ssrt} & 69.90 & 58.90 & 66.00 & & 75.80 & 59.80 & 60.20 & & 73.20 & 70.60 & 62.20 & & 71.40 & 61.70 & 55.20 & 65.41 \\
         MemSAC \cite{memsac} & 63.49 & 42.14 & 60.32 & & 72.33 & 54.92 & 46.14 & & 73.46 & 68.04 & 52.75 & & 74.42 & 57.79 & 43.57 & 59.11 \\
         CDTrans \cite{transformer_da} & 66.20 & 52.90 & 61.50 & & 72.60 & 58.10 & 57.20 & & 72.50 & 69.00 & 59.00 & & 72.10 & 62.90 & 53.90 & 63.16 \\
         PMTrans \cite{pmtrans} & 74.10 & 61.10 & \textbf{70.00} & & 79.30 & 63.70 & 62.70 & & 77.50 & 73.80 &  62.60 & &  79.80 &  69.70 &  61.20 &  69.63\\ \midrule
         \multicolumn{14}{l}{\textit{\textbf{Language-guided UDA}}} \\
         TextMatch \cite{lagtran} & 71.36 & 64.30 & 65.32 & & 81.25 & 65.65 & 64.85 & & 81.09 & 72.65 & 63.94 & & 81.08 & 70.84 & \textbf{64.17} & 70.14 \\
         nGramMatch \cite{lagtran} &  68.92 & 59.82 & 63.15 & & 76.35 & 61.72 & 62.87 & & 76.35 & 69.28 & 62.51 & & 76.04 & 68.52 & 60.52 & 67.17 \\
         CLIP+TextMatch \cite{lagtran} & 72.89 & 63.56 & 66.97 & & \underline{81.53} & 62.98 & 66.74 & & \textbf{81.89} & 72.81 & 65.97 & & \textbf{81.87} & 72.07 & 63.34 & 71.05 \\
         LaGTran \cite{lagtran} & \underline{77.30} & \underline{68.25} & 67.35 & & 81.31 & \underline{67.03} & 66.81 & & 80.78 & \underline{75.62} & \underline{68.08} & & 79.23 & \underline{73.80} & 63.44 & \underline{72.41} \\

         \midrule
         \textbf{TRUST} & \textbf{81.04} & \textbf{71.12} & \underline{69.95} & & 81.42 & \textbf{69.15} & \textbf{69.03} & & 81.44 & \textbf{79.37} & \textbf{72.77} & & 81.61 & \textbf{78.33} & \underline{64.11} & \textbf{74.95}\\
        \bottomrule
    \end{tabular}
    \end{adjustbox}
    \caption{Classification accuracy (\%) under the classical shifts of DomainNet-345. All methods use the Swin-base backbone. The symbol * indicates models finetuned on the source data.}
    \label{tab:domainnet_results}
    \vspace{-1em}
\end{table*}

\begin{table}
    \centering
    \begin{adjustbox}{width=\columnwidth}
    \begin{tabular}{lccccc}
        \toprule
         &  \multicolumn{2}{c}{GeoImnet}  &\multicolumn{2}{c}{GeoPlaces} &\multirow{ 2}{*}{\shortstack{Total \\ Avg.}}\\
         \cline{2-3} \cline{4-5}
         & U $\rightarrow$ A & A $\rightarrow$ U & U $\rightarrow$ A & A $\rightarrow$ U &\\
         \midrule
         \multicolumn{5}{l}{\textit{\textbf{Zero-shot classification}}} \\
         CLIP \cite{clip} & 49.84 & 53.83 & 43.41 & 54.34  & 50.36 \\
         CLIP* \cite{clip} & 57.79 & 59.12 & 48.91 & 55.89 & 55.42 \\
         \midrule
        \multicolumn{5}{l}{\textit{\textbf{UDA}}}\\
         Source only & 52.46 & 51.91 & 44.90 & 36.85 &  46.53 \\
         CDAN \cite{cdan} &  54.48 & 53.87 & 42.88 & 36.21 &  46.86 \\
         MemSAC \cite{memsac} &  53.02 & 54.37 & 42.05 & 38.33 &  46.94 \\
         ToAlign \cite{toalign} &  55.67 & 55.92 & 42.32 & 38.40 &  48.08 \\
         MDD \cite{mdd} & 51.57 & 50.73 & 42.54 & 39.23 &  46.02\\
         DALN \cite{daln} & 55.36 & 55.77 & 41.06 & 40.41 &  48.15\\
         PMTrans \cite{pmtrans} & 56.76 & 57.60 & 46.18 & 40.33 &  50.22\\ \midrule
        \multicolumn{5}{l}{\textit{\textbf{Language-guided UDA}}} \\
         TextMatch \cite{lagtran} & 49.68 & 54.82 & 53.06 & 50.11 & 51.92 \\
         nGramMatch \cite{lagtran} & 49.53 & 51.02 & 51.70 & 49.87 &  50.93 \\
         CLIP+TextMatch & 50.11 & 54.92 & 50.36 & 52.14 &  51.88 \\
         LaGTran \cite{lagtran} & \underline{63.67} & \underline{64.16}  & \underline{56.14} & \underline{57.02} & \underline{60.24}\\
         \midrule
         \textbf{TRUST} & \textbf{65.77} & \textbf{67.02}  & \textbf{59.89} & \textbf{60.14} & \textbf{63.20} \\
        \bottomrule
    \end{tabular}
    \end{adjustbox}
    \caption{Classification accuracy (\%) under the geographical shifts of GeoNet. All methods use a ViT-base backbone.} 
    \label{tab:geonet_results}
    \vspace{-1em}
\end{table}

\subsection{Overall Framework}
\label{subsection:overall_framework}
To adapt the source to the target domain, we train a vision model on both the source and target domains, combining the source labels and the target pseudo-labels obtained by the finetuned BERT model in \cref{subsec:lagtran}.
Differently than \cite{lagtran}, we reweight the classification loss for target samples based on the reliability of the pseudo-labels estimated through CLIP scores in \cref{subsec:uncertainty}. The higher the estimated reliability, the more it will contribute to the classification loss. Hence, we train \name~with the following classification losses:

\begin{equation}
    \mathcal{L}^{cls}_s = \frac{1}{|\mathcal{D}_s|} \sum_{i=1}^{|\mathcal{D}_s|} \mathcal{L}_{\operatorname{CE}} (h(f(a_w(x^i_s))), y^i_s),
\end{equation}
\begin{equation}
\begin{split}
    \mathcal{L}^{cls}_t = \frac{1}{|\mathcal{D}_t|} \sum_{i=1}^{|\mathcal{D}_t|} &w_i  \cdot \mathcal{L}_{\operatorname{CE}}(h(f(a_w(x^i_t))), \tilde{y}_t^i) + \\ 
    &(1-w_i)  \cdot  \mathcal{L}_{\operatorname{CE}}(h(f(a_w(x^i_t))), p_t^i), 
\end{split}
\end{equation}

\noindent where $\mathcal{L}_{\operatorname{CE}}$ is the cross-entropy loss, $w_i$ is the reliability weight calculated from CLIP (\textit{c.f.} \Cref{subsec:uncertainty}), $p_t^i=h(f(a_s(x^i_t)))$ are predictions obtained by the vision model on strongly augmented samples, and $a_w$ and $a_s$ are weak and strong augmentations, respectively.
Finally, on target images and captions, we employ our proposed soft-contrastive framework optimising $\mathcal{L}^{ctr}$. 

The overall loss function is the following:

\begin{equation}
    \mathcal{L} = \mathcal{L}^{cls}_s + \mathcal{L}^{cls}_t + \mathcal{L}^{ctr}.
\end{equation}

\section{Experimental Results}
\label{sec:results}
Datasets, implementation details and additional analyses are reported in the supplementary material.

\noindent \textbf{Baselines}. We employ two text baselines (TextMatch and nGramMatch) as in \cite{lagtran} to evaluate the effect of using the source data for fine-tuning the language model, instead of using directly the target captions. We also compare with CLIP employing two baselines: CLIP zero-shot inference \cite{language_da}, performed by incorporating the domain information into the text prompt of CLIP (\textit{e.g.}, \emph{A clipart of a} \texttt{<class>}); CLIP model finetuned on the source to adapt it to the data distribution. Finally, we employ a training-free baseline by ensembling CLIP and TextMatch (CLIP+TextMatch) by averaging the CLIP's image-text similarity and the TextMatch's caption-label similarity, and then computing the argmax to obtain the ensemble predictions.

\subsection{Results}
\label{subsec:results}
\textbf{DomainNet-345}. \Cref{tab:domainnet_results} presents results on DomainNet-345 that, exhibiting classical shifts, leads to generally higher scores than in  GeoNet. CLIP and CLIP* achieve a notable accuracy of $70.38\%$ and $71.26\%$. \name~ outperforms standard UDA approaches, CLIP and CLIP* by $+5.32\%$, $4.57\%$ and $+3.69\%$, respectively. On non-real to real transfers, \name~ performs worse than CLIP and CLIP+TextMatch. A possible explanation is that CLIP is trained on a massive amount of real-world data, potentially eliminating any domain shifts between the train and test settings. Differently, \name~ is trained on non-real images (the source domain), which instead exhibit domain shifts with testing data. Nonetheless, \name~ achieves overall superior performance across diverse target domains, demonstrating its adaptability to different domain shifts. Compared to LaGTran, which also uses text during training, \name~ improves accuracy on average by $+2.54\%$, demonstrating the effectiveness of the introduced components. Moreover, \name~ performs best in almost all the transfer scenarios, achieving the best  or  second best results on $9$ out of $12$ scenarios. \\
\textbf{GeoNet}. \Cref{tab:geonet_results} presents results of \name~on geographical shifts of GeoNet. Our method outperforms the source-only baseline by $+16.67\%$. Compared to LaGTran \cite{lagtran}, we improve performance by $+2.96\%$, achieving better performance on all the shift settings of both GeoImnet and GeoPlaces. We also compare our method with the zero-shot performance of CLIP and CLIP+TextMatch. Despite CLIP being trained on a significantly larger amount of data, our approach obtains a gain in performance of $+12.84\%$. While in most of the settings the CLIP+TextMatch performs better than CLIP and TextMatch used alone, results are still lower than \name~. We hypothesise that CLIP and TextMatch make coherent mistakes, leading to a low improvement when ensembling the two models. Finally, compared to the CLIP model finetuned on source data (CLIP*), \name~ improves accuracy by $+7.78\%$. \\ 
\textbf{VisDA}. Although in \name,  CLIP is not finetuned and used only for uncertainty estimation, we compare with other approaches that use CLIP for UDA and we present results on VisDA in \cref{tab:uda_visda}. \name~ outperforms PADCLIP by $+2.1\%$ on average, achieving the best performance on 4 out of 12 classes while maintaining competitive accuracies on the other classes (see Tab. 1 in the supplementary material). These results also demonstrate the effectiveness of \name~ on large-scale datasets.

\subsection{Analysis}
\label{subsec:analysis}

\begin{table}[t!]
    \centering
    \begin{adjustbox}{width=0.7\columnwidth}
    \begin{tabular}{lc}
        \toprule
        Method  &  Avg. Acc. \\
        \midrule
        DAPL \cite{dapl} & 86.9 \\
        DAMP \cite{damp} & 88.4 \\
        UniMoS \cite{unimos} &  88.1 \\
        PADCLIP \cite{padclip} & 88.5 \\
        \midrule
        \textbf{TRUST} & \textbf{90.6}\\

        \bottomrule
    \end{tabular}
    \end{adjustbox}
    \caption{UDA performance (\%) on VisDA dataset with ResNet101 compared with CLIP-based approaches. (Full table in Supplementary Material.)}
    \label{tab:uda_visda}
    \vspace{-1em}
\end{table}



\begin{figure}[t!]
    \centering
    \includegraphics[width=0.5\columnwidth]{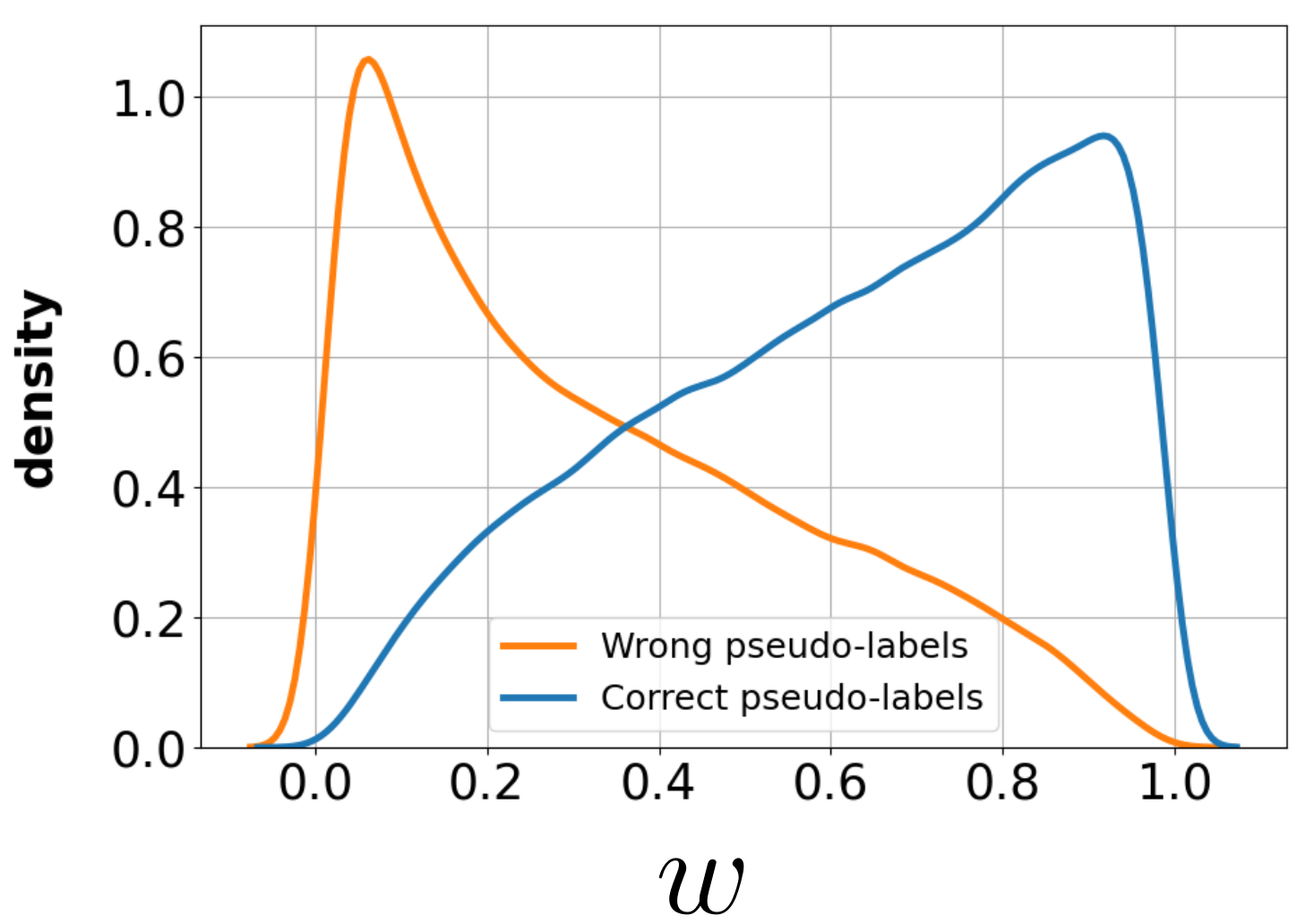}
    \caption{Probability density function of reliability weights estimated in \cref{subsec:uncertainty} for target samples having correct and wrong pseudo-labels.} 
    \label{fig:clip_weights}
    \vspace{-1em}
\end{figure}

\textbf{Ablation Studies}. In \Cref{tab:ablation} (Left), we report ablation studies for the \name~components on GeoNet. When using only  the pseudo-labels, \name~achieves the lowest accuracy of $62.31\%$. When adding the standard contrastive loss \cite{simclr} (second row), the improvement in performance remains negligible. But when using our proposed language guided soft-contrastive loss (\cref{subsection:soft_contrastive}), we boost the performance by $+2.52\%$ (third row). The fourth row presents the results enabling only our CLIP-based uncertainty estimation (\cref{subsec:uncertainty}), which brings a gain in performance of $+1.10\%$. Finally, in the last row, we show the gain obtained by the full \name~model which, enabling both the two introduced components, further improves performance by $+4.09\%$.\\ 
\noindent \textbf{Generating pseudo-labels with CLIP.}
\Cref{tab:additional_analyses} (Right) presents results  using CLIP \cite{language_da} for generating pseudo-labels in \cref{subsec:lagtran}, as an alternative to finetuning BERT. We compare three different CLIP-based approaches: the CLIP zero-shot inference (TRUST w/ CLIP), the CLIP model finetuned on source data (TRUST w/ CLIP*) and the generation of pseudo-labels by evaluating the similarity between captions and class names in the CLIP's text feature space (TRUST w/ CLIP-Text). Results show that using CLIP (instead of BERT) for obtaining the pseudo-labels  leads to lower performance. Moreover, text-only based approaches, like CLIP-Text and BERT, achieve better results, justifying the assumption that language models benefit of a larger robustness to domain shift than vision models.\\
\noindent \textbf{Effectiveness of CLIP-based uncertainty estimation.}
\cref{fig:clip_weights} illustrates the effectiveness of the proposed CLIP-based uncertainty estimation strategy, showing the distribution of the reliability weights $\mathbf{w}$ (\cref{subsec:uncertainty}) for target samples with correct and wrong pseudo-labels $\mathbf{\title{y}_t}$. \cref{fig:clip_weights} shows that the two distributions are clearly separable and samples with correct pseudo-labels have high values of $\mathbf{w}$ (high reliability), while samples with wrong pseudo-labels have lower values of $\mathbf{w}$ (low reliability). These results validate our strategy to use CLIP to identify low-quality text descriptions, which is able to down-weight wrong pseudo-labels and to reduce their contribution in the classification loss.

\begin{figure}[t!]
    \centering
    \includegraphics[trim=5cm 0 0 0, clip,width=.8\columnwidth]{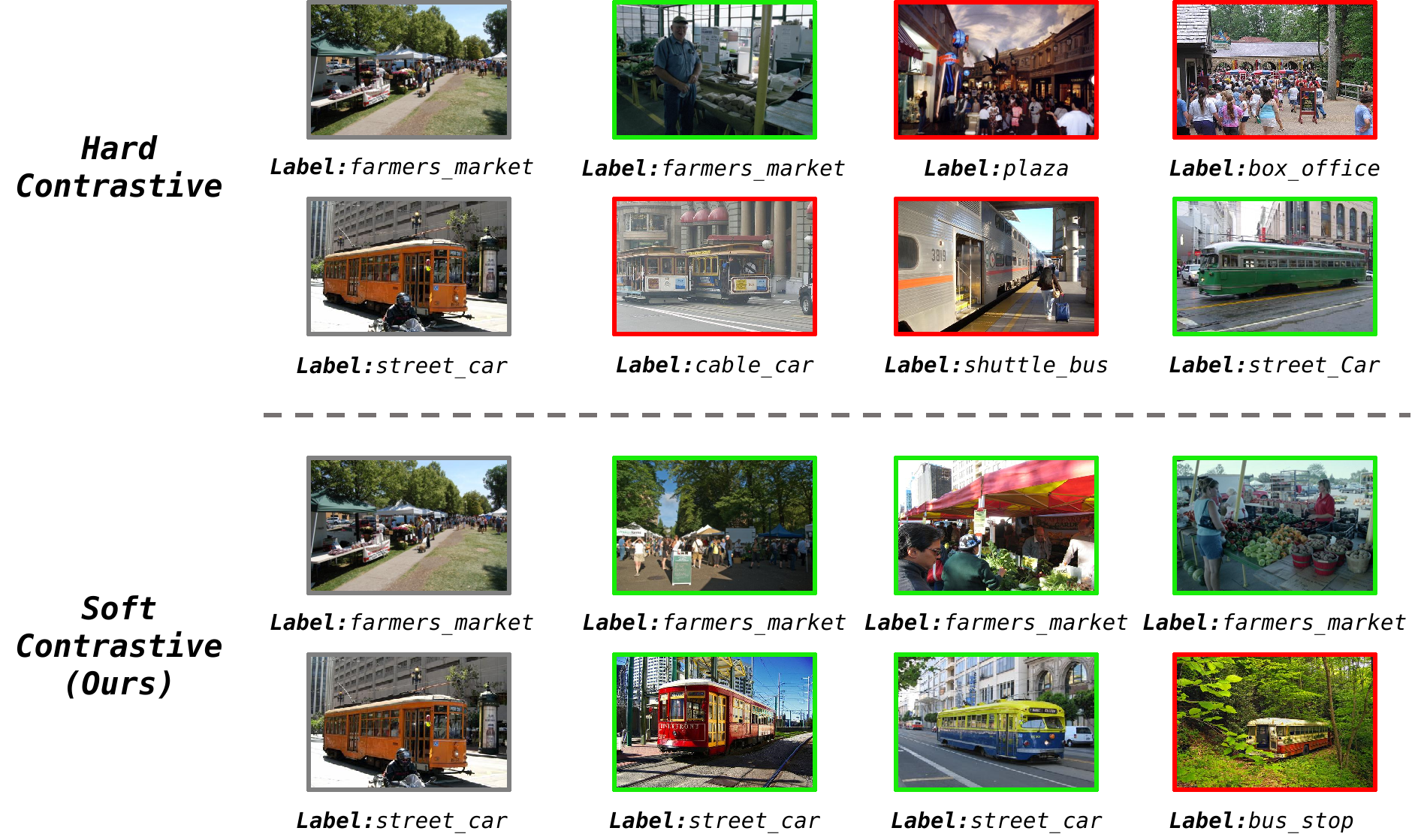}
    \caption{Visualisation of nearest neighbors for 2 target images (grey borders) comparing the standard contrastive loss \cite{simclr} (top 2 rows), and our soft-contrastive loss (\textit{c.f.} \cref{subsection:soft_contrastive}) (bottom 2 rows). Images with green/red borders represent correctly/incorrectly retrieved images.} 
    \label{fig:retrieval}
    \vspace{-1.5em}
\end{figure}

\begin{table}[t!]
\centering
\begin{minipage}{0.48\linewidth}
\centering
\begin{adjustbox}{width=\linewidth}
\begin{tabular}{c|ccc}
\toprule
\shortstack{Hard \\ Contrastive} & \shortstack{Soft \\ Contrastive} & \shortstack{CLIP \\ Uncertainty} & \shortstack{Avg. \\ Acc.}
\\ \midrule
\xmark   &  \xmark  &  \xmark   & 62.31 \\
\cmark & \xmark &  \xmark  & 62.86 \\
\xmark & \cmark & \xmark & 65.38 \\
\xmark & \xmark & \cmark & 63.41 \\
\xmark & \cmark & \cmark & \textbf{66.40} \\
\bottomrule
\end{tabular}
\end{adjustbox}
\end{minipage}
\hfill
\begin{minipage}{0.48\linewidth}
\centering
\begin{adjustbox}{width=\linewidth}
\begin{tabular}{lc}
\toprule
Method & Avg. Acc. \\
\midrule
TRUST w/ CLIP & 61.36\\
TRUST w/ CLIP* & 63.32\\
TRUST w/ CLIP-Text & 64.11\\
\textbf{TRUST w/ BERT*} & \textbf{66.40}\\
\bottomrule
\end{tabular}
\end{adjustbox}
\end{minipage}
\caption{(Left) Ablation studies of components of the proposed method. (Right) Classification accuracy (\%) on GeoImnet comparing CLIP and BERT for generating pseudo-labels (\cref{subsec:lagtran}). The symbol * indicates models finetuned on the source data.}
\label{tab:additional_analyses}
\label{tab:ablation}
\vspace{-1em}
\end{table}
\noindent \textbf{Effects of the soft-contrastive loss.}
\cref{fig:retrieval} shows a qualitatively analysis of the effectiveness of our soft-contrastive loss in aggregating the representations of semantically similar samples that belong to the same class. We show the top-3 nearest neighbour retrievals using features computed by the vision model, when trained with the standard hard contrastive loss \cite{moco,simclr}, or with the proposed soft-contrastive loss. Our loss produces a more fine-grained retrieval even in presence of semantically similar classes (e.g. \emph{streetcar}, \emph{cable\_car}, \emph{shuttle\_bus}). On the contrary,  the hard contrastive loss leads to less accurate retrievals, likely due to a coarser-grained feature space, where visually similar classes are aggregated even if they represent different semantic concepts (e.g. \emph{farmers\_market} and \emph{plaza}).

\section{Conclusions}
\label{sec:conclusions}
In this work, we introduced \name, a novel approach to UDA for image classification that leverages textual data to guide the adaptation to the target domain. \name~ generates pseudo-labels for target samples from captions and estimates their uncertainty using CLIP to mitigate the impact of wrong pseudo-labels in the classification loss. We also proposed a novel soft-contrastive learning framework that aligns vision and language feature space, to transfer the shift robustness from the language to the vision model. Our extensive evaluations on DomainNet and GeoNet benchmarks demonstrated that \name~ outperforms the current state-of-the-art in both classical and complex domain shifts.


\bibliography{main}


\end{document}